\documentclass[conference]{IEEEtran}
\IEEEoverridecommandlockouts

\usepackage{cite}
\usepackage{amsmath,amssymb,amsfonts}
\usepackage{graphicx}
\usepackage{textcomp}
\usepackage[table]{xcolor}
\def\BibTeX{{\rm B\kern-.05em{\sc i\kern-.025em b}\kern-.08em
    T\kern-.1667em\lower.7ex\hbox{E}\kern-.125emX}}
\usepackage{cleveref}
\usepackage{subcaption}
\usepackage{caption}
\usepackage[linesnumbered,ruled]{algorithm2e}
\usepackage{xpatch}
\usepackage{multirow}
\usepackage{multicol}
\usepackage{nomencl}
\usepackage{siunitx}
\usepackage{blindtext}
\usepackage[T1]{fontenc}
\usepackage[utf8]{inputenc}
\usepackage[spaces,hyphens]{xurl}
\usepackage{float}
\usepackage{enumitem}

\usepackage[left=.75in,right=.75in,top=1.09in,bottom=1.24in]{geometry}

\SetKwComment{Comment}{$\triangleright$\ }{}
\SetKwInput{kwInit}{Input}
\SetKwInput{KwResult}{Output}

\usepackage{fancyhdr}
\fancypagestyle{mystyle}{
    \chead{\large The 24th IEEE International Conference on Mobile Data Management (MDM 2023)}
}

\newcommand{\fref}[1]{Fig.~\ref{#1}}

\newcommand{\sref}[1]{Section~\ref{#1}}

\newcommand*{\Resize}[2]{\resizebox{#1}{!}{$#2$}}%

\newcommand\HUGE{\fontsize{21.91}{30}\selectfont}
\begin{document}
\title{\HUGE One-Shot Federated Learning for LEO Constellations that Reduces Convergence Time from Days to 90 Minutes}
%

\author{\IEEEauthorblockN{Mohamed Elmahallawy}
\IEEEauthorblockA{\textit{Department of Computer Science} \\
\textit{Missouri University of Science and Technology}\\
Rolla, MO 65409, USA\\
meqxk@mst.edu}
\and
\IEEEauthorblockN{Tie Luo$^*$}\thanks{$^*$Corresponding author. This work is supported by the National Science Foundation under Grant No. 2008878.}
\IEEEauthorblockA{\textit{Department of Computer Science} \\
\textit{Missouri University of Science and Technology}\\
Rolla, MO 65409, USA\\
tluo@mst.edu}
}
\maketitle
\thispagestyle{mystyle}

\begin{abstract}

A Low Earth orbit (LEO) satellite constellation consists of a large number of small satellites traveling in space with high mobility and collecting vast amounts of mobility data such as cloud movement for weather forecast, large herds of animals migrating across geo-regions, spreading of forest fires, and aircraft tracking.
Machine learning can be utilized to analyze these mobility data to address global challenges, and Federated Learning (FL) is a promising approach because it eliminates the need for transmitting raw data and hence is both bandwidth and privacy-friendly. However, FL requires many communication rounds between clients (satellites) and the parameter server ($\mathcal {PS}$), leading to substantial delays of up to several days in LEO constellations. In this paper, we propose a novel {\em one-shot} FL approach for LEO satellites, called {\em LEOShot}, that needs only {\em a single communication round} to complete the entire learning process. LEOShot comprises three processes: (i) {\em synthetic data generation}, 
(ii) {\em knowledge distillation}, 
and (iii) {\em virtual model retraining}. 
We evaluate and benchmark LEOShot against the state of the art and the results show that it drastically expedites FL convergence by {\em more than an order of magnitude}. Also surprisingly, despite the one-shot nature, its model accuracy is on par with or even outperforms regular iterative FL schemes by a large margin.
\end{abstract}


\section{Introduction}\label{sec:intro}

Low Earth orbit (LEO) satellite constellations have recently been burgeoning due to the rapid advances in satellite communications (SatCom) technology. Positioned at an altitude of 160--2,000 km above the Earth's surface, LEO satellites are often equipped with sensors and high-resolution cameras to collect a vast amount of mobility-related data, such as tracking and monitoring cloud movements for weather forecast \cite{perez2021airborne}, hurricane and forest fire movements \cite{barmpoutis2020review}, flooding situations, migration of large herds of animals across geographic regions, and aircraft tracking. Large-scale machine learning (ML) models can be utilized to analyze these mobility data to address global challenges such as climate change, natural disasters, and abnormal wildlife conditions. However, traditional ML methods require downloading raw data such as satellite images to a ground station (GS) or gateway for centralized model training, which is not practical for SatCom because of its limited bandwidth, large propagation delay, and privacy concerns (e.g., satellite data and images may contain sensitive information such as military activities or critical infrastructure locations).



Introducing federated learning (FL) \cite{fl2021} to SatCom appears to be a viable solution because FL eliminates the need for transmitting raw data by allowing satellites to train ML models locally (i.e., on-board) using their own data respectively and only send the resulting model parameters to the GS which acts as the $\mathcal {PS}$ to aggregate those local models into a global model. However, FL requires many communication rounds between clients (satellites) and the $\mathcal {PS}$ to re-train and re-aggregate the models until it converges into a well-functioning global model. As a result, this iterative process can take several days or even weeks in the context of SatCom \cite{so2022fedspace,razmischeduling}, because of the long propagation delay and, most importantly, the highly {\em sporadic and irregular connectivity} between LEO satellites and the GS. The latter is attributed to the distinct trajectories of satellites and the GS,\footnote{A satellite orbits at an {\em inclination angle} between 0$^o$ and 90$^o$ (typically 50-80$^o$), whereas a GS constantly rotates on the 0$^o$ plane. These degrees are in reference to the Equator of the Earth.} which leads to very intermittent and non-cyclic visits of satellites to the GS (successively).

In this paper, we propose LEOShot, a novel {\em one-shot} FL approach for LEO satellite constellations that accomplishes the FL training process in a single communication round, yet still obtaining a global model with competitive performance. One-shot FL is an emerging paradigm~\cite{oneshot2021jmlr} but its existing methods cannot be directly applied to SatCom because they (i) require the $\mathcal {PS}$ to have a publicly shareable dataset that represents the client data distribution \cite{guha2019one, li2020practical}, which is hardly available, (ii) require clients to upload raw or synthetic data \cite{kasturi2020fusion,zhou2020distilled}, which contradicts the FL principle, or (iii) still needs extra communication rounds to achieve a satisfactory accuracy \cite{zhang2021practical}. In contrast, LEOShot does not share or transmit data in any form (e.g., raw or synthetic) yet still attains a high-performing model in just a single communication round. In summary, this paper makes the following contributions:
\begin{itemize}[leftmargin=*]
\item To the best of our knowledge, LEOShot is the first one-shot FL approach proposed for SatCom. It drastically reduces the adverse effect of highly sporadic and irregular connectivity between satellites and GS by instating only one communication round. Not only this, but it also operates without relying on any auxiliary datasets or raw-data sharing, yet attaining competitive model performance.

\item LEOShot comprises (i) {\em synthetic data generation} 
that generates images mimicking real satellite images instead of downloading them; (ii) {\em knowledge distillation} that trains (instead of aggregates) a global model using a teacher-student framework; iii) {\em virtual model retraining} that refines the global model iteratively toward high performance but without any extra communication rounds. 

\item Unlike conventional FL which assumes an identical ML model architecture for all clients, LEOShot allows clients to use {\em different} model architectures to cater to their own preferences and resource constraints. This is much more flexible and helps solve the {\em data heterogeneity} and {\em system heterogeneity} among LEO satellites.

\item We demonstrate via extensive simulations that LEOShot dramatically accelerates FL convergence by {\em more than an order of magnitude} as compared to the state-of-the-art FL-SatCom approaches. In the meantime, the accuracy of its trained models outperforms existing methods by a large margin despite its one-shot nature.
\end{itemize}

\noindent\textbf{Paper Organization.} \sref{sec:releated} provides an overview of recent work that utilizes the FL in SatCom. \sref{sec:model} describes the FL-SatCom network model as well as communication links among satellites and GS. \sref{LEOShot_model} explains the proposed LEOShot framework and its component in detail. The performance evaluation of LEOShot is provided in \sref{evaluation}. Finally, \sref{sec:conc} concludes this paper.

\section{Related Work} \label{sec:releated}

\subsection{FL for SatCom}
While FL-SatCom is still in its infancy, a few studies have attempted to apply FL to SatCom. Chen et al. \cite{chen2022satellite} directly applied FedAvg \cite{mcmahan2017communication} to SatCom to demonstrate FL's advantages in avoiding the need to download raw data to a GS. FedISL \cite{razmi} leverages intra-plane inter-satellite-link (ISL) to reduce the long waiting time for satellites to become visible to the $\mathcal {PS}$, 
but only performs well under an ideal setup where the $\mathcal {PS}$ is either a GS located at the North Pole (NP) or a medium Earth orbit (MEO) satellite positioned directly above the Equator. In addition, FedISL overlooks the {\em Doppler shift}  resulting  from the high relative speed between LEO satellites (clients) and MEO-satellite ($\mathcal {PS}$). Another approach called FedHAP \cite{happaper} was proposed which introduces one or multiple high altitude platforms (HAPs) floating at 18-25km above the Earth's surface as $\mathcal {PS}$. As a result of using HAPs, the convergence speed of FL is improved by having more visible satellites, but more hardware is required. Another most recent approach, FedLEO \cite{e2023opt}, has been proposed to enhance the convergence process of FL in the LEO constellation. FedLEO  achieved this through the use of intra-plane model propagation and scheduling of sink satellites. It is however necessary for each satellite to run a scheduler to determine which satellite will be the sink to send its model, resulting in a delay for models to be exchanged with the GS.
 
The above are all {\em synchronous FL} approaches. There are also {\em asynchronous FL} approaches proposed for SatCom that allow the $\mathcal {PS}$ to proceed to the next training rounds without waiting for the model updates from {\em all} the clients. Razmi et al. \cite{razmi2022ground} proposed FedSat, which assumes that the GS is located at NP to simplify the problem so that every satellite visits the GS at {\em regular intervals} (once every orbital period). To overcome this limitation, they proposed another approach FedSatSchedule \cite{razmischeduling}, which uses a scheduler to reduce model staleness by predicting satellites' visiting patterns to the GS. However, several days are required to reach a model convergence. Another recent approach called FedSpace  \cite{so2022fedspace} formulated an optimization problem to dynamically schedule model aggregation based on predicting connectivity between LEO satellites and the GS, but it requires each satellite to upload a small portion of its data so that the GS can schedule model aggregation, which contradicts FL's principle of avoiding raw data sharing. Another recently proposed FL-SatCom approach, AsyncFLEO \cite{mAsyFLEO}, is proposed to offer a drastic solution to the staleness challenge that requires several days for asynchronous FL-SatCom approaches to converge. It is capable of achieving convergence within a few hours, but is still subject to a high number of communication rounds and hence there is still large room for improvement.

\subsection{One-Shot FL}
To the best of our knowledge, one-shot FL has not been introduced into SatCom before. The studies \cite{guha2019one, li2020practical} employ knowledge distillation \cite{gou2021knowledge} in which client-side models are used as teacher models to train a server-side student model (global model). However, the server is required to have access to a public dataset in order to provide pseudo samples for training purposes, which conflicts with FL's privacy principles. 
As an alternative to knowledge distillation, data distillation \cite{wang2018dataset} is used in \cite{zhou2020distilled, kasturi2020fusion}. These studies, however, show a notable underperformance; moreover, they require uploading distilled/synthetic data generated by clients to a $\mathcal {PS}$, which is incompatible with FL principles. Lastly, a theoretical analysis of one-shot FL is presented in \cite{salehkaleybar2021one}, focusing on independently and identically distributed (IID) data only.

Given the above challenges, none of the existing one-shot frameworks can be practically applied without issues or drawbacks. In addition, to develop a one-shot FL approach for SatCom, it is necessary to take into account the SatCom challenges such as long propagation delays, intermittent and irregular visibility of LEO satellites, and bandwidth limitations.

\section{Network Model} \label{sec:model}

\begin{figure}[t]
     \centering
     \includegraphics[width=\linewidth]{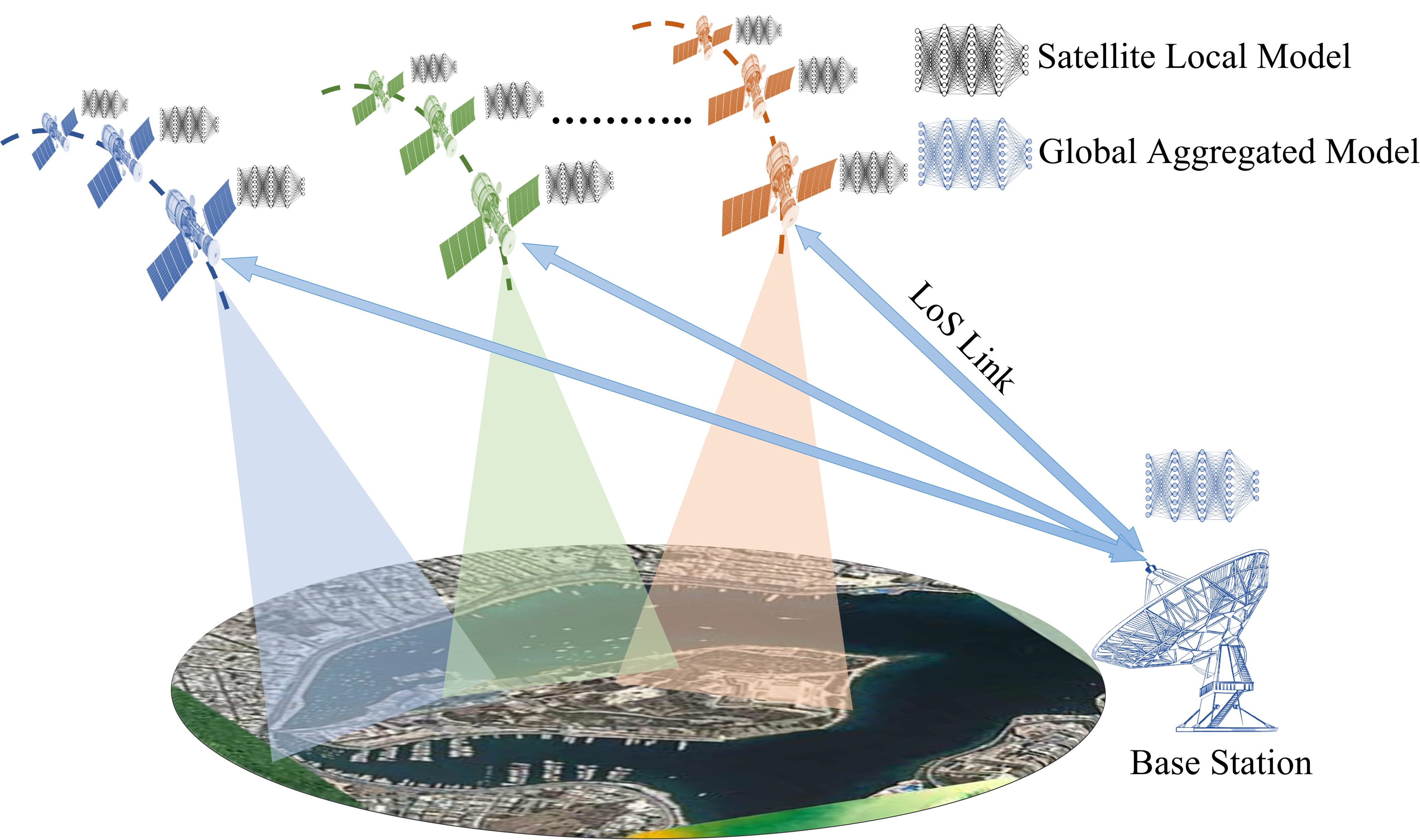}
    \caption{FL-LEO network architecture, comprised of multiple orbits each having multiple satellites.}
    \label{system_model}
\end{figure}

\fref{system_model} illustrates an LEO constellation $\mathcal M$ consists of $N$ satellites equally distributed on $O$ orbits. Each orbit $o \in \mathcal{O}=\{o_1,o_2,...,o_O\}$ is located at an altitude $h_{o}$ above the Earth with an inclination angle $\alpha_{o}$ and has a set of satellites $\mathcal{M}_{o}$. Satellites in an orbit $o$ travel with the same velocity $v_{o}$ and has the same orbital period $T_{o}$. Here, $v_{o}=\sqrt{\frac{GM_{E}}{(R_{E}+h_{o})}}$ and $T_{o}= \frac{2 \pi}{\sqrt{GM_{E}}}{(R_{E}+h_{o})^{3/2}}$, where $G$ is the gravitational constant, $M_{E}$ is the Earth's mass, and $R_{E}=6,371$ km is the Earth's radius. 

In SatCom, satellite $m$ can communicate with a GS $g$ if the Earth does not obstruct their line-of-sight (LoS) link. Mathematically, this means $\vartheta_{m,g}(t) \triangleq \angle (r_{g}(t), (r_{m}(t) - r_{g}(t))) \leq \frac{\pi}{2}-\vartheta_{min}$, where $r_{m}(t)$ and $r_{g}(t)$ are the trajectories of $m$ and $g$, respectively, and $\vartheta_{min}$ is the minimum elevation angle that depends on the GS location.

\subsection{FL-SatCom System}

Consider an FL-SatCom system, in which each satellite $m\in\mathcal M$ collects a set of Earth images  $\mathcal {D}_{m}$ of size $d_{m}=|\mathcal {D}_{m}|$. In addition, we assume that these images are non-IID among satellites since they orbit the Earth at irregular intervals and scan different areas. In a synchronous FL system such as FedAvg \cite{mcmahan2017communication}, the $\mathcal {PS}$ (i.e., a GS) and satellites train an ML model collaboratively to solve the following problem
\begin{equation} \label{eqn1}
           \arg \min_{w \in \mathbb{R}^{d}} F(w)= \sum_{m\in \mathcal M }{\frac{d_{m}}{d}{F_{m}{(w)}},}
\end{equation}
where $F(w)$ indicates the overall loss function (e.g., SSE) of the target model; $w$ is the model weights; $d= \sum_{m\in \mathcal M} d_{m}$ is the total data size; $F_{m}(w)$ is the loss function of satellite $m$, which can be expressed as
\begin{equation}
     F_{m}{(w)} = \frac{1}{d_{m}}\sum_{j=1}^{d_{m}} f_{m}(w; x_{m,j}),
\end{equation}
where $f_{m}(\cdot)$ is the training loss on a sample point $x_{m,j}$.

During the training process, there are multiple communication rounds $\beta =\{0, 1, 2,\dots\}$. In each round, the GS first transmits the latest global model weights $w^\beta$ to all satellites, and each satellite $m$ employs a local optimization scheme such as gradient descent to update its model for $I$ epochs as
\begin{equation}
    w_{m}^{\beta,i+1} = w_{m}^{\beta,i}- \eta \nabla F_{m}(w_{m}^{\beta,i}; x_{m}^{i}), ~ i=0,1,2,...,I-1
\end{equation}
where 
$\eta$ is the learning rate. Following that, each satellite uploads its locally updated model to the GS for assembling as
\begin{equation}
    w^{\beta+1} = \sum_{m\in \mathcal M} \frac{d_{m}}{d} w_{m}^{\beta,I},
\end{equation}
which completes a communication round. The above procedure iterates with an incremental $\beta$ until the FL model converges (e.g., a target accuracy, target loss, or the maximum number of training rounds is reached).

{\bf A key challenge} of this learning process arises from the fact that the convergence of FL requires several communication rounds between LEO satellites and the GS, and each communication can only happen when a satellite transiently comes into the GS's visible zone. As a result, FL would take several days or even longer to reach convergence. This motivated us to develop LEOShot, which requires only a single communication round between satellites and the GS.
\begin{figure*}[h!]
     \centering
          \subfloat[{\bf Phase 1:} Synthetic data generation.]
         {\label{data}
        \includegraphics[width=1\textwidth]{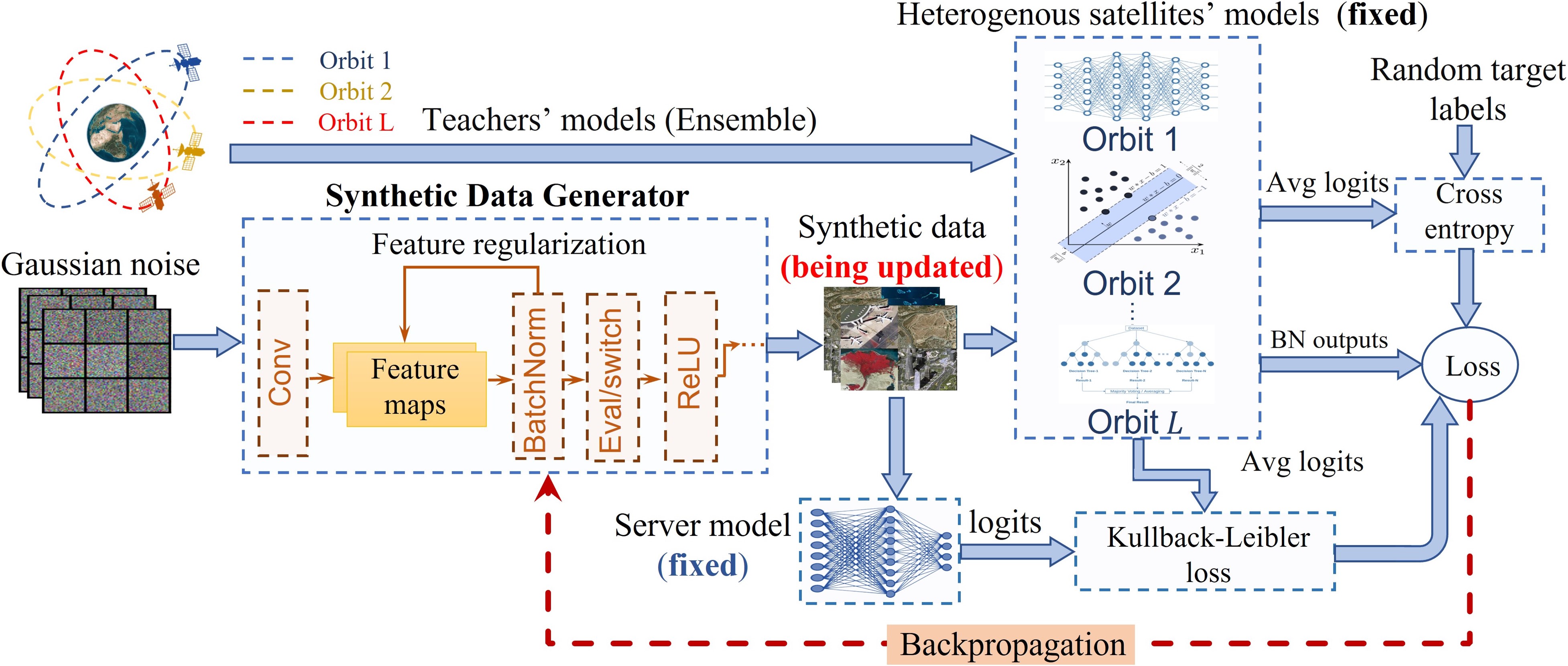}}\\ \vspace{0.4cm}
          \subfloat[{\bf Phase 2:} Knowledge distillation.]
         {\label{distillation}
        \includegraphics[width=0.56\textwidth]{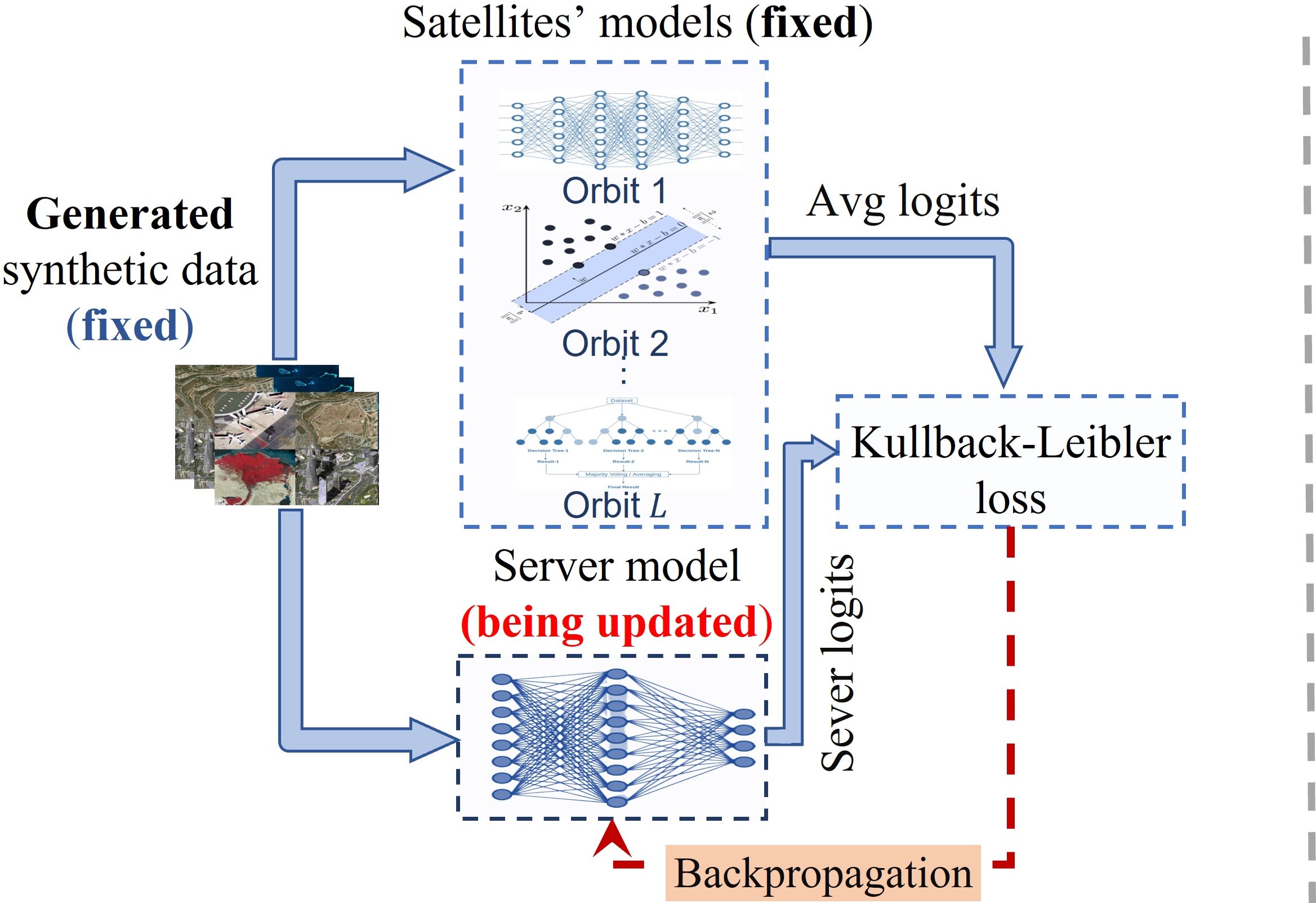}}
       \subfloat[{\bf Phase 3:} Virtual model retraining.]
         {\label{Convergence} \includegraphics[width=0.44\textwidth]{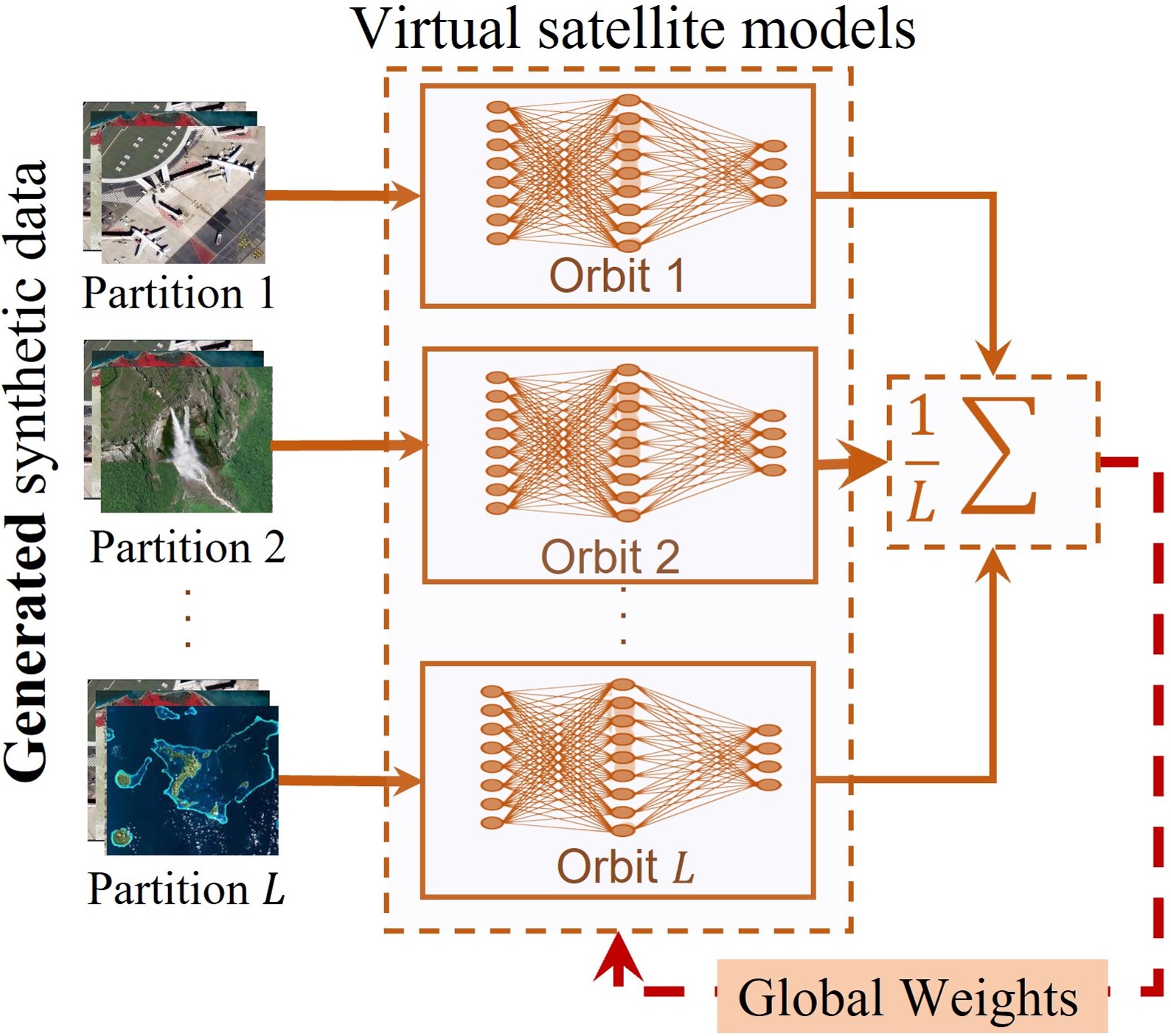}}
        \caption{Overview of the proposed LEOShot framework.}
        \label{fig:synthetic_data}
\end{figure*}

 \subsection{Communication Model}\label{Com_link}

In a symmetric radio frequency channel with additive white Gaussian noise (AWGN), the signal-to-noise ratio (SNR) between a satellite $m$ and a GS $g$ can be expressed as \cite{leyva2021inter}:
\begin{equation}
    SNR(m,g) = \frac{P G_{m}G_{g}}{K T B {L}_{m,g}}, 
\end{equation}
where $P$ is the transmitter power, $G_{m}$ and $G_{g}$ are the total antenna gains of satellite $m$ and an GS $g$, respectively, $K$ is the Boltzmann constant ($1.38\times10^-23 J/K$), 
$T$ is the noise temperature at the receiver, $B$ is the channel bandwidth, and ${L}_{m,g}$ is the free-space pass-loss between satellite $m$ and a GS $g$. As long as the LoS link between a satellite $m$ and a GS $g$ is not obstructed by the Earth, then ${L}_{m,g}$ can be expressed as 
\begin{equation}
    {L}_{m,g} = \bigg(\frac{4\pi \|m,g\|_{2}  f}{c}\bigg)^{2}  
\end{equation}
where $\|m,g\|_{2}$ is the Euclidean distance between a satellite $m$ and a GS $g$ when they are visible to each other, $f$ is the carrier frequency, and $c$ is the light speed. For exchanging local or global model weights ($w_m~\text{or}~w$) between a satellite $m$ and a GS $g$, the total required time $t_{t}$ can be calculated as
\begin{equation}
     t_{t} = {\underbrace{\frac{z{|\bf{\mathcal{P}|}}}{R}}_{\text{\bf Transmission delay}}+\underbrace{\frac{\|m,g\|_{2}}{c}}_{\bf \text{\bf Propagation delay}}}+t_{m}+ t_{s} \label{eqny}
\end{equation}
where $t_{m}$ and $t_{s}$ are the processing delay at {\em m-th} satellite and a GS $g$, respectively, ${|\mathcal{P}|}$ is the number of sample points, $z$ is the number of bits in each sample, and $R$ is the maximal achievable data rate, which can be computed by the Shannon formula as
\begin{equation} \label{eqnz}
    R \approx B \log_{2} (1+SNR(m,g))
\end{equation}

\section{The LEOShot framework}\label{LEOShot_model}


LEOShot is operated at the server end. \fref{fig:synthetic_data} provides an overview of the LEOShot framework, which is comprised of three phases: (a) synthetic data generation, which synthesizes a representative dataset as a {\em proxy} of all the satellites' local data, (b) knowledge distillation, for distilling information from the satellite models (teacher models) to train a server model (student model), and (c) virtual model retraining, for training virtual local models iteratively until the server model converges. Algorithm~\ref{algorithm1} outlines the entire process of LEOShot.

\begin{algorithm}
\caption{LEOShot's 3-phase process}\label{algorithm1}
\kwInit{Satellites' models, $L,\eta_g,\eta_s$, $x$, $y$, $J$,$I$ ,$\gamma_1, \gamma_2$}
\KwResult{Server model parameters}
 \Comment*[h]{{\bf Phase 1:} Synthetic data generation }\\
  Generate batches of random noises $x$ and labels $y$\\
  Initialize server model (student) and generator $\mathcal G$\\
  \ForEach{epoch $j$ of training $\mathcal G$}{
   Calculate losses  $\mathcal R_{CE},\mathcal R_{BN},{R}_{{KL}_{Gen}}(\hat x) $ using eqns. \eqref{Average_logits}, \eqref{BatchNorm}, and \eqref{gen_loss}\\
   Calculate the generator loss $\mathcal R'_{Gen}$ via \eqref{final_generator}\\
   Generate/Update $\hat x$\\
   Retrain the weights of $\mathcal G$ on generated $\hat x$ via \eqref{final_generator}\\
  }
\Comment*[h]{{\bf Phase 2:} Knowledge distillation}\\
{
  \ForEach{epoch $j$ of training server model}{
   Calculate loss $\mathcal R_{{KL}_S}$  of the server model via \eqref{server_loss}\\
   Retrain the weights of the server model via \eqref{server_model}\\
  }
  {\bf return} $w_{s}$\\
}
\Comment*[h]{ {\bf Phase 3:} Virtual Model Retraining}\\{
Generate $L$ virtual models\\
Cluster $\hat x$ into $L$ Partitions\\
\ForEach{epoch $j$ of updating server model }{
\ForEach{virtual model $w_l$}{ \Comment{\footnotesize All models train in parallel}
Initialize $w_l \gets w_{s}$\\
Train $w_l$ using its assigned partition of $\hat x $  \\
}
Update $w_{s} \leftarrow \frac{1}{L} \sum_{l=1}^{L} w_{l}$
}
{\bf return} $w_{s}$
}
\end{algorithm}

As a background, the server begins its operation once it receives all the client models from all the orbits via {\em sink satellites}. A sink satellite \cite{happaper} is a satellite on each orbit who (1) collects all the models from other satellites on the same orbit (via intra-orbit model relay), (2) assembles them together into a {\it partial global model} \cite{happaper} and (3) sends it to the server.

\subsection{Synthetic Data Generation}\label{sec:synth_gen}

The objective of this phase is to build a generator $\mathcal G$ to generate high-quality unlabeled synthetic data (i.e., having sufficient features for correct classification) without having to download any real data from satellites to a GS or requiring any auxiliary (i.e., publicly available) datasets. To achieve this objective, we use the ensemble of the client (satellite) models uploaded by various sink satellites to generate synthetic data (see \fref{data}). Note that a salient feature of LEOShot is that it allows {\em heterogeneous model architectures} of these client models. That is, each client can decide and use its own preferred neural network rather than a common architecture used by all the satellites (as dictated by the standard FL). We achieve this by using an {\em ensemble} of all the client models and {\em train} the target global model via {\em knowledge distillation} (instead of by averaging model weights).

Our generator $\mathcal G$ was inspired by the generative adversarial network (GAN) \cite{lin2020ensemble}, but is distinct from it as we do not require public datasets. Given a randomly generated input $x$ (e.g., Gaussian white noise) and a random label $y$, our generator attempts to generate synthesized data $\hat x$ with similar features to the data collected by satellites, by solving the following optimization problem:
\begin{equation}\label{optimization}
     \min_{\hat x} \mathcal R_{Gen} \triangleq \mathcal{R}_{CE}(D(\hat x), y) +\mathcal R(\hat x),
\end{equation}
where $\mathcal R_{Gen}$ denotes the overall loss of the generator, $\mathcal{R}_{CE}(\cdot)$ is a cross-entropy loss (e.g., classification loss), and $\mathcal R(\hat x)$ is a regularization term used to steer $\hat x$ towards more realistic data (e.g., images). Here, $D(\hat x)$ is the average logits of the ensemble model given an input $\hat x$ (where logits are the output of usually the last fully connected layer), and is defined by
\begin{equation} 
    D(\hat x)= \frac{1}{|\mathcal M|} \sum_{l\in \mathcal L} f_{l}(\hat x, w_{\mathcal K_l}), 
    \label{Average_logits}
\end{equation}
where $f_{l}(\hat x, w_{\mathcal K_l})$ is an estimation function of the ensemble model received from orbit $l$, that returns the logits of $\hat x$ when the model parameter $w_{\mathcal K_l}$ is given. Eq. \eqref{Average_logits} allows us to (indirectly) measure how well the generated synthetic data $\hat x$ mimics both the distribution and the particular instances of the original satellite data, without accessing the original data. In fact, attempting to access satellites' training data contradicts the FL principles. Moreover, unlike traditional FL algorithms (e.g., FedAvg), our design of generator \eqref{optimization} uses logits (as in $D(\cdot)$) instead of client model weights, which enables our approach to deal with {\em heterogeneous} client models. 


The regularizer $\mathcal R(\hat x)$ serves the purpose of improving the stability of the generator. The need comes from the fact that the ensemble of satellite models is trained on non-IID datasets, and therefore tend to make the generator unstable and get stuck in suboptimal local minima. Additionally, there is a risk of overfitting the synthetic data, which ultimately hinders the generator from achieving high levels of accuracy \cite{li2021fedbn}. 
Therefore, we use a BatchNorm (BN) layer during training to normalize the feature maps to reduce the impact of covariate shifts \cite{ioffe2015batch} and overcome the gradient vanishing problem (so that the distance between synthetic images and original satellite images can be continuously reduced). In addition, this BN layer also implicitly stores the average logits as channel-wise means and variances. Thus finally, the $\mathcal R(\hat x)$ realized by this BN layer can be expressed by
\begin{align}\label{BatchNorm}
\Resize{7.7cm}{\mathcal {R}_{BN}(\hat x)= \frac{1}{L} \sum\limits_{l\in \mathcal L} \sum\limits_{b}\Bigl({\|\mu_b(\hat x)-\mu_{l,b}\|}_{2}+ {\|\sigma_{b}^{2}(\hat x)-\sigma_{l,b}^{2}\|}_{2}\Bigl)}
\end{align}
where $\mu_b(\hat x)~\text{and}~\sigma_{b}^{2}(\hat x)$ are the batch-wise estimation of the mean and variance, respectively, associated with the \textit{b-th} BN layer of the generator, and $\mu_{l,b}~\text{and}~\sigma_{l,b}^{2}$ are the mean and variance of the \textit{b-th} BN layer of $f_{l}(\hat x, w_{\mathcal K_l})$. As a result of adding this regularization term, our designed generator outputs synthetic images with high quality that are close to the original satellite images.

\subsection{Knowledge Distillation }\label{sec:know_tran}

In this phase, we strive to distill the knowledge from the ensemble of all the client models to train a server (i.e., global) model. Although the generated synthetic data has high quality, it is not useful for knowledge distillation  because of the large gap between the ensemble model's decision boundaries and the server model's decision boundary \cite{heo2019knowledge}. To address this issue, we force the generator to generate more synthetic data with different distributions and then employ Kullback-Leibler (KL) divergence to minimize the distance between the predictions (proxy of decision boundary) of the ensemble model and the server model during synthetic data generation (bottom of Fig.~\ref{data}). The new regularizer term that represents the KL divergence is
\begin{align}\label{gen_loss}
\Resize{1.7cm} {\mathcal {R}_{{KL}_{Gen}}(\hat x)} &=\Resize{5.55cm} { 1- \frac{1}{2} \bigl\{KL{(D_{G}(\hat x), Q)}+KL(D_{E}(\hat x), Q)) \bigl\}} \\
Q &=\frac{1}{2} \cdot (D_{G}(\hat x)+D_{E}(\hat x)),
\end{align}
where $KL(\cdot)$ denotes the KL divergence loss, $D_{G}(\hat x)$ is the server model's logits, and $D_{E}(\hat x)$ is the ensemble model's average logits. 

Thus, with our BN and KL regularization terms, we reformulate our generator optimization problem \eqref{optimization} as
\begin{equation}\label{final_generator} 
\Resize{7.7cm}{\mathcal R'_{Gen}=  \min_{\hat x}
 { \Big[\mathcal R_{CE}(D(\hat x), y) +\gamma_1 {\mathcal R}_{BN}(\hat x)+ \gamma_2 {R}_{{KL}_{Gen}}(\hat x) \Big]  }}
\end{equation}
where $\gamma_1~\text{and}~\gamma_2$ are hyper-parameters to trade-off between the two losses. In addition, we optimize the weights of our generator $\mathcal G$ for $J$ epochs using SGD as follows:
\begin{align}
\Resize{7.7cm}{w_{Gen}^{j} = w_{Gen}^{j-1}- \eta_g \nabla \mathcal R_{Gen}{(w_{Gen}^{j-1}; {(\hat x, y)^{j-1})}, \quad j=1,2,\dots,J}}
    \label{generator_eqn}
\end{align}
where $w_{Gen}^{j}$ is the generator's model at iteration $j$, and $\eta_g$ is the $\mathcal G$'s learning rate.

After these optimization procedures, our generator can generate synthetic images not only of high quality but also of a distribution that resembles the original satellite data to enable effective knowledge distillation.

Referring to Fig.~\ref{fig:synthetic_data}b, the next step after generating the synthetic data $\hat x$ is to commence the updating and retraining of the server model until it attains an acceptable accuracy. To this end, the synthetic data $\hat x$ is fed to both the ensemble of satellite models which acts as the teacher, and the server model which acts as the student. Subsequently, the average logits are computed as the outcome of training the teacher model using \eqref{Average_logits}, which can be used for both homogeneous and heterogeneous ensemble models. The average logits are then applied to distill the knowledge from the ensemble (teacher) model to the server (student) model by minimizing the prediction error between the two models, through the KL divergence as follows:
\begin{align}\label{server_loss}
\mathcal {R}_{{KL}_{S}}(\hat x) = KL(D(\hat x), D_{s}(\hat x)) 
\end{align}
where $D_s(\hat x)$ is the logits of the server model after being trained on the generated synthetic data $\hat x$. 
Since the satellite models are trained on non-IID data, we aim to improve the accuracy of the server model and address the issue of poor performance or divergence problem encountered in \cite{li2021fedbn}. To achieve this, we further optimize the server model parameters by employing SGD as
\begin{align}
\Resize{7.7cm}{w_{s}^{j} = w_{s}^{j-1}- \eta_s \nabla \mathcal R_{{KL}_S}{(w_{s}^{j-1}; {\hat x^{j-1})}, \quad j=1,2,\dots,J}}
    \label{server_model}
\end{align}
where $w_{s}^{j}$ is the server model at iteration $j$, and $\eta_s$ denotes the learning rate of the server model. Note that our method is different from \cite{li2021fedbn} which uses local batch normalization at each client to harmonize local data distributions but requires several rounds of communication between server and clients. Instead, we use generated synthetic data to resemble the original satellite data and it allows us to directly train a server model locally using SGD, without the need for communication or averaging satellite models which are influenced by non-IID data distributions. On the downside, synthetic data may not be as good as real data; to address this, in Phase 3 we retrain our server model to improve model accuracy.


As a result of this phase, we have successfully trained a server model that leverages the knowledge from the ensemble satellite models and the generated synthetic data, and we have taken into account the possible heterogeneity of both the data and models involved.
\fref{distillation} illustrates the entire knowledge distillation process. 

\subsection{Virtual Model Retraining}\label{sec:retrain}

Although the knowledge distillation  phase outputs a functional server model, the model performance still has notable room for improvement (e.g., the classification accuracy was merely 70\% on the MNIST dataset). Certainly, allowing for extra communication rounds (and aggregation) between the GS and the LEO satellites will improve model accuracy, but this clearly contradicts our goal of one-shot learning and will also negatively impact the convergence speed.
Thus, we propose a novel method that transforms the distributed FL into a {\em localized} version, by creating {\em virtual local models} on the server and trains those models {\em locally} until the server model converges. Specifically, our method consists of four steps:
(i) clone $L$ copies of the server model to serve as the initial virtual local models, (ii) partition the generated synthetic data (after labeling them using a K-means clustering algorithm) into $L$ groups, each with the same class distribution as one of the $L$ orbits (distributions were received at the end of model dissemination), (iii) train each virtual model on one of the $L$ data groups, (iv) aggregate the weights of these trained virtual models to obtain an updated server model. The above repeats until the server model converges, which is the final global model. \fref{Convergence} illustrates the entire retraining process for virtual models. 




\section{Performance Evaluation}\label{evaluation}
\subsection{Simulation setup}
\begin{table}[b!]

\setlength{\tabcolsep}{1.3em}
\centering
\renewcommand{\arraystretch}{1.3}
\caption{Simulation Parameters (upper: communication; lower: training)}
\label{Parameter}
\begin{tabular}{|l|l|}
 \hline
  \centering \textbf{Parameters} & \textbf{Values} \\
   \hline \hline
  Transmission power (satellite \& GS) $P$& 40 dBm \\
  \hline
  Antenna gain of (satellite \& GS) $G_m , G_{s}$& 6.98 dBi\\
  \hline
 Carrier frequency $f$ & 2.4 GHz \\
  \hline 
 Noise temperature $T$& 354.81 K \\
  \hline 
 Transmission data rate $R$ & 16 Mb/sec   \\ 
 \hline \hline  
   Number of local training epochs $I$ & 300 \\
  \hline 
   Learning rate $\eta$ & 0.001 \\
  \hline 
   Mini-batch size $b_k$& 32\\
  \hline 
   Generator learning rate $\eta_g$ & 0.001 \\
\hline 
  Weighting factors $\gamma_1 \& \gamma_2$ & 1 \& 10\\
\hline 
\end{tabular}
\end{table}
\textbf{LEO Constellation.} We consider a Walker-delta constellation $\mathcal M$, which consists of 40 LEO satellites distributed over five orbits, each with eight satellites. Each orbit is located at an altitude $h_{o}$ of 500km above the Earth's surface with an inclination angle of 80$^\circ$. A GS is located in Rolla, MO, USA (can be anywhere) with a minimum elevation angle $\vartheta_{min}$ of 10$^\circ$. For both LEOShot and baselines, Table \ref{Parameter} (upper part) summarizes the parameters pertaining to the communication links described in \sref{Com_link}. By using Systems Tool Kit (STK), a software tool for analyzing satellite constellations, we extract the visibility between satellites and the GS. To obtain each set of results, we simulate communication between satellites and the GS over a period of three days.
\begin{figure*}[h]
\centering
         {\includegraphics[width=0.8\textwidth]{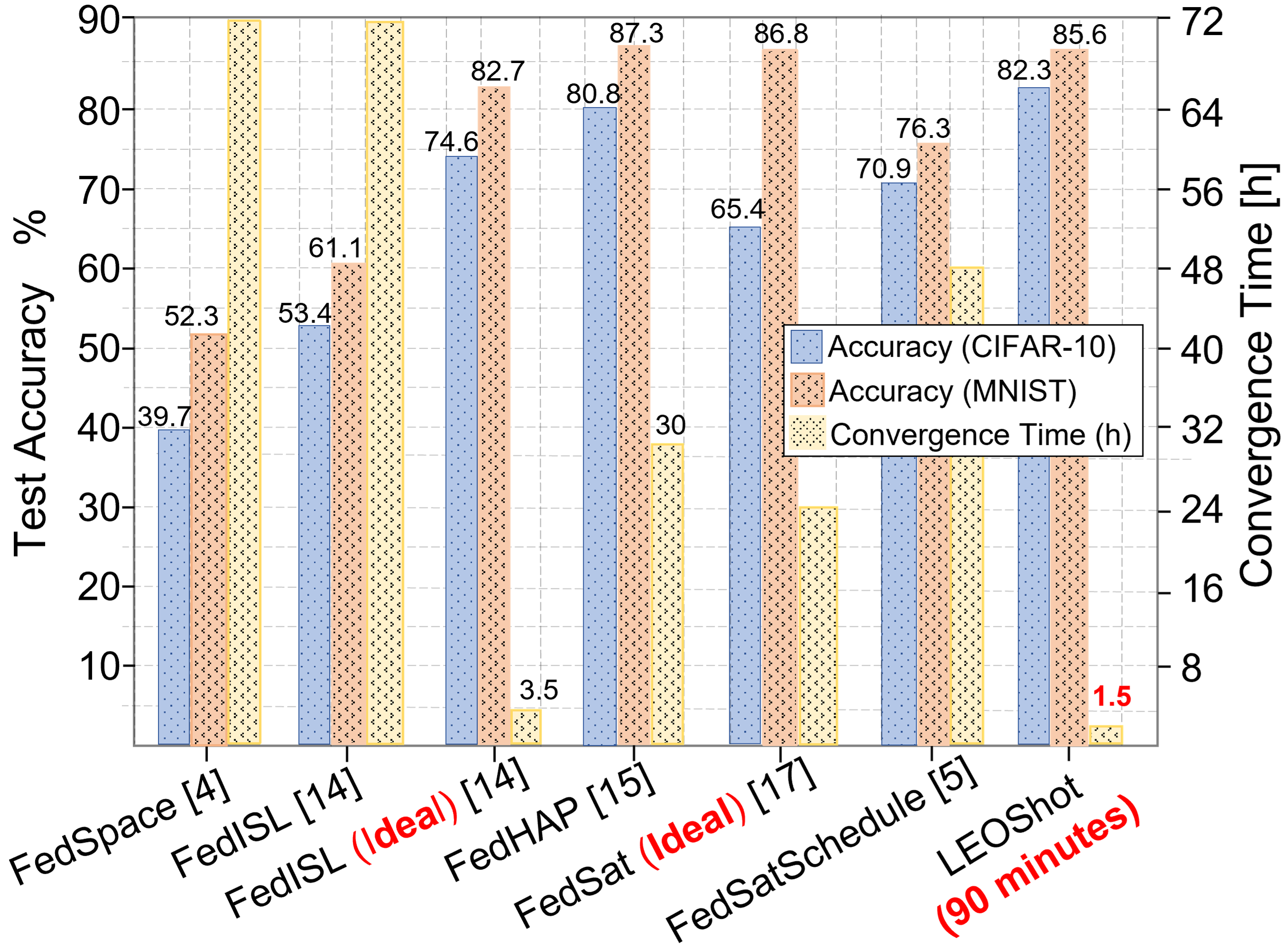}}
        \caption{Accuracy and Convergence Time comparison on non-IID data. For all approaches, convergence time was first measured on MNIST and then fixed for CIFAR-10 to measure the accuracy.} \label{benchmark}

\end{figure*}
\textbf{Baselines.} We compare LEOShot with the state-of-the-art approaches that were proposed most recently and are reviewed in \sref{sec:releated}, including
FedSpace~\cite{so2022fedspace},
FedISL~\cite{razmi}, FedHAP~\cite{happaper},  FedSat~\cite{razmi2022ground}, and FedSatSchedule \cite{razmischeduling}. 

\textbf{ML models and Dataset.} An important implication of LEOShot's capability of allowing heterogeneous client models (cf. \sref{sec:synth_gen}), is that the server no longer needs to broadcast an initial model $w^0$ to all the clients, which in the context of SatCom will save a significant amount of time. However, for comparison with existing methods, we assume a standard (homogeneous) FL setting where the neural network architecture is common and broadcasting $w^0$ is still required. 
{\em We highlight that this setup substantially favors baseline approaches} and not ours. In the experiment, all satellites train a ResNet-50 model. For each baseline, the GS aggregates the satellite models into a global ResNet-50 model. For LEOShot, however, since a key component of knowledge distillation  is to  train a smaller global model, the GS trains a ResNet-18 model. 

For comparison purposes, we use the same datasets as all the baseline approaches use: MNIST \cite{deng2012mnist}  and CIFAR-10 \cite{CIFAR-10}. 
Additionally, we consider a non-IID setting for both datasets, where satellites in two orbits are trained with 4 classes while satellites in the other three orbits are trained with the remaining 6 classes. The lower part of Table \ref{Parameter} summarizes the training hyperparameters.

\subsection{Results}
{\bf Comparison with Baselines.} As shown in \fref{benchmark}, LEOShot achieves the fastest convergence (on both datasets) in only {\bf 90 minutes} with an accuracy of 85.64\% on MNIST attained in a single communication round (the convergence time includes waiting for at least one satellite per orbit to be visible to upload a partial model to the GS). The second fastest approach is FedISL \cite{razmi} with the {\em ideal} setup described in \sref{sec:releated} (GS at NP or MEO above the Equator), which takes 3.5 hours for FL to converge with an accuracy of 82.67\%. Without the ideal setup, its convergence time spikes to 72 hours, and accuracy drops to 61.19\%. FedSat \cite{razmi2022ground} and FedHAP \cite{happaper} have marginally higher accuracy than LEOShot but their convergence is significantly slower. Also importantly, FedSat assumes an ideal setup (GS at NP) similar to FedISL, and FedHAP requires extra hardware (HAP) of substantial extra cost. FedSatSchedule \cite{razmischeduling} does not have FedSat's ideal assumption, and as a result, its accuracy is only 76.32\% and its convergence time doubles FedSat.

For all methods, accuracy on CIFAR-10 is lower than on MNIST after the same amount of training time, which is particularly prominent for FedSat \cite{razmi2022ground}. On the other hand, LEOShot maintains a very small difference, which demonstrates certain robustness. Recall that LEOShot achieves high accuracy in just a {\em single communication round}.

\begin{figure}[h]
\centering
    \subfloat[Samples of generated synthetic images (MNIST dataset).]
    { \includegraphics[width=0.45\textwidth]{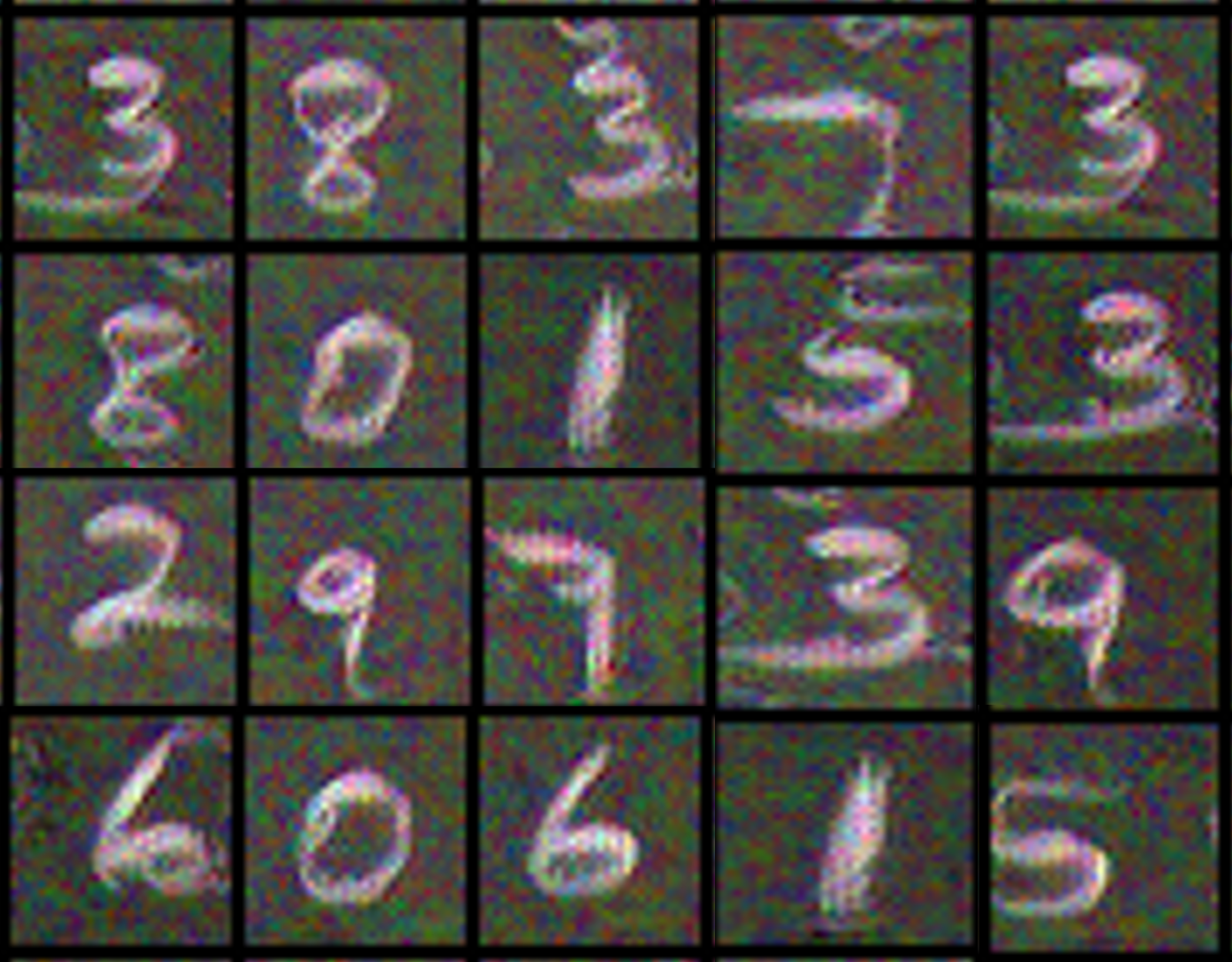}}\vfill\vspace{3mm}
    \subfloat[Samples of generated synthetic images (CIFAR-10 dataset).]
{ \includegraphics[width=0.45\textwidth,height=6cm]{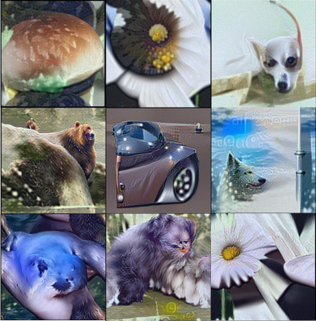}}
{\caption{Samples of synthetic images created by our generator when satellites trained on various datasets. \label{Sample}}} 
\end{figure}

{\bf Effectiveness of using synthetic data in knowledge distillation.} Pertaining to \sref{sec:synth_gen} and \ref{sec:know_tran}, this subsection (i) investigates whether using model {\em logits} or model {\em parameters} (weights or gradients) is more effective in producing a server model, and (ii) assesses the test accuracy on multiple deep learning (DL) models with the generated synthetic data. Fig. \ref{Sample} shows a sample of the generated synthetic images using logits from satellites trained on MNIST and CIFAR-10 datasets. These images approximate the distribution and mimic the content of the real images very well, enabling an effective knowledge distillation.

\begin{table}[ht]
\setlength{\tabcolsep}{1.3em}
\renewcommand{\arraystretch}{1.3}
\caption{Comparison of DL models trained on synthetic images}
\label{table13}
\centering
 \begin{tabular}{|p{3.2cm}|p{1.2cm} | p{1.7cm}|}
 \hline
 DL model &\multicolumn{2}{c |} {Accuracy (\%)}\\
 \cline{2-3}
 &\rmfamily MNIST& CIFAR-10 \\
 \hline 
 CNN (2 layers) & 62.26 & 60.88\\
 \hline 
 VGG-11& 69.15&62.72 \\
  \hline 
 Wide-ResNet-40-1 \cite{zagoruyko2016wide}& 71.51&66.79 \\
 \hline 
    \rowcolor{gray!20}
  \textbf{ResNet-18} &  \textbf{73.64}& \textbf{70.67} \\
 \hline 
\end{tabular}
\end{table}

\begin{figure}[h]
\centering
    {\includegraphics[width=\linewidth]{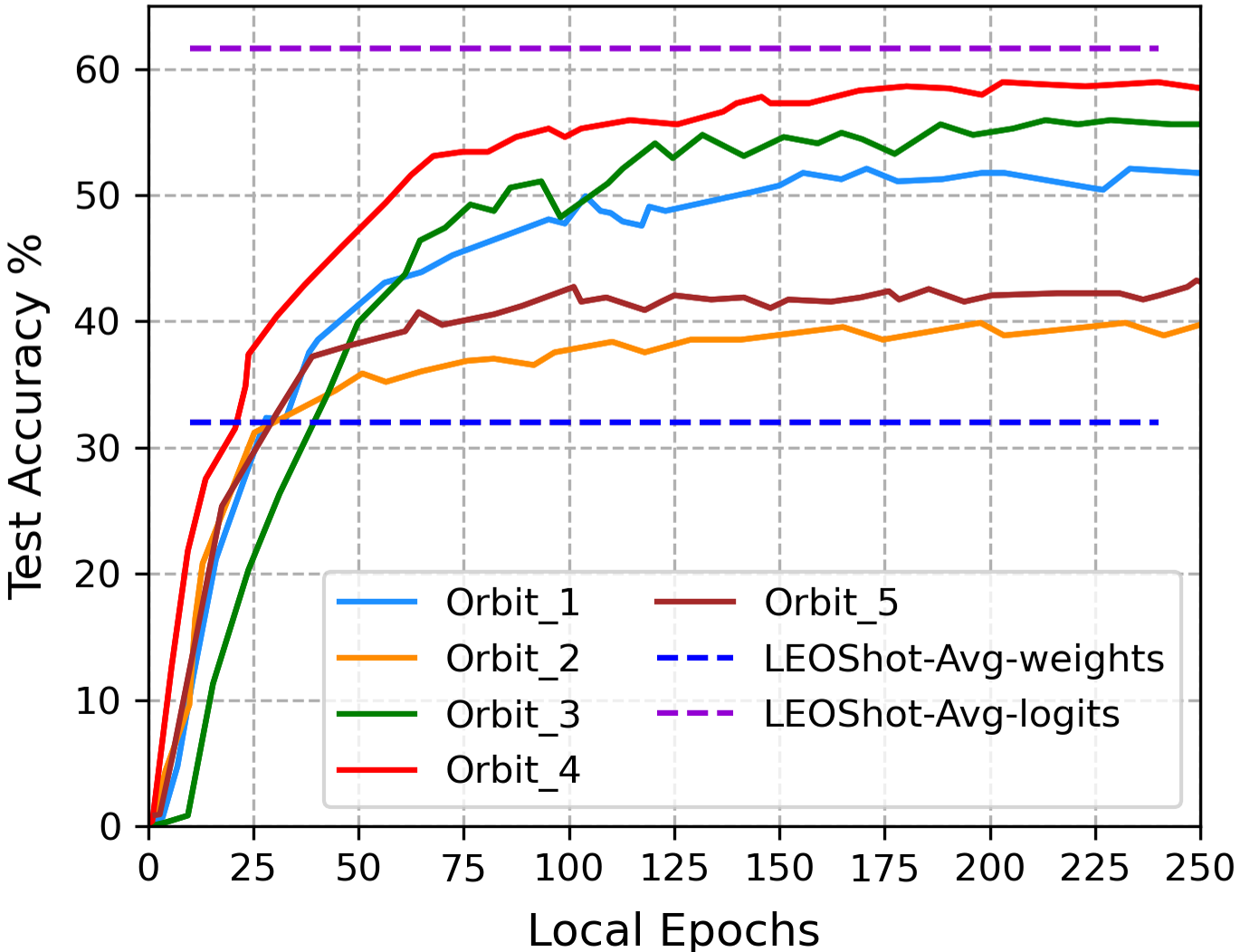}}
        \caption{Test accuracy of partial models from sink satellites on different orbits (solid lines) and the best accuracy obtained by averaging logits or weights (dotted lines).} \label{Training_models}
\end{figure}

For the first purpose, we conducted two experiments on LEOShot. In the first experiment, satellite model parameters are uploaded to the GS similarly to traditional FL. In the second experiment, model logits are uploaded instead. All the satellites train a ResNet-50 model on MNIST in the non-IID setting. Fig.~\ref{Training_models} shows the resulting test accuracy of each partial global model derived from each orbit, as well as the accuracy after averaging the weights or logits. The results indicate that the test accuracy of each partial global model varies depending on the data imbalance from each orbit. The accuracy of the server model is only 31.74\% when averaging these partial models' parameters, which underperforms the individual partial models. In contrast, when averaging the logits, the server model achieves an accuracy of 62.36\%, which outperforms all the individual partial models. This interesting finding confirms that a single communication round is far from sufficient for weight averaging (as in traditional FL) to accommodate the discrepancy between client models trained on non-IID data; but on the other hand, averaging logits allows knowledge distillation through our {\em local} training that minimizes the distance between logits of our teacher and student models. 
In addition, using logits also allows us to accommodate model heterogeneity.


For the second purpose, Table~\ref{table13} reports the performance of various DL models when trained using our synthesized images. We can see that all the DL models achieve acceptable accuracy, with ResNet-18 achieving the highest 73.64\% on MNIST. This set of results validates that the quality of our synthesized images is reliable, which plays an important role in transferring knowledge to the server model.

{\bf Impact of virtual model retraining.} Pertaining to \sref{sec:retrain}, here we investigate how our use of virtual local models affects the accuracy of the server model. As reported in Table~\ref{table13}, the server model achieves an accuracy of 73.64\% by using ResNet-18 as the target model and MNIST as the training dataset. Note that this is achieved without virtual model retraining. Now, with that, we clone five virtual ResNet-18 models initialized with the initial server model weights and train each virtual model on one of five partitions of synthetic images. The final server model is obtained via weight averaging. Our result in \fref{benchmark} shows that the server model accuracy increases to 85.64\% which is a significant 16\% improvement without requiring any extra communication round.


\section{Conclusion and Future work} \label{sec:conc}

This work makes the first effort to introduce one-shot FL into SatCom. We propose a novel framework called LEOShot to address the challenge of highly sporadic and irregular visits of LEO satellites to a GS. Unlike prior work, LEOShot does not require public datasets or client data uploads, thus upholding important FL principles on data privacy protection and communication efficiency. In addition, unlike standard FL which dictates a common identical neural network architecture for all the clients and the server, LEOShot allows each client to choose its own preferred ML model based on its computing resources and data properties. In our quantitative study in comparison with the state-of-the-art benchmarks, we find that LEOShot reduces FL training/convergence time drastically up to 80 times (it converges in as short as 90 minutes); in the meantime, it achieves high accuracy even under challenging non-IID settings and outperforms the benchmarks by large margins.

In our future work, we aim to examine LEOShot on real and diverse satellite datasets in different settings. This would include exploring a variety of LEO constellations ranging from sparse to dense constellations with GS located at different geographical locations, as well as training heterogeneous ML models across satellites and constellations.


\bibliographystyle{IEEEtran}
\large \bibliography{references.bib}

\end{document}